\pdfoutput=1
\documentclass[11pt]{article}
\usepackage[final]{acl}
\usepackage{times}
\usepackage{latexsym}
\usepackage[T1]{fontenc}
\usepackage[utf8]{inputenc}
\usepackage{microtype}
\usepackage{inconsolata}
\usepackage{graphicx}
\usepackage{tabularx}
\usepackage{subfigure}
\usepackage{subcaption}
\usepackage{amsmath}
\usepackage{multirow}
\usepackage{float}
\usepackage{xcolor}
\usepackage{booktabs}
\usepackage{makecell}
\usepackage[colorinlistoftodos]{todonotes}

\title{\newGispyName: a Novel Metric for Biomedical Text Simplification \\ via Gist Inference Score}

\author{Chen Lyu \\
  University of Warwick \\
  \texttt{chen.lyu@warwick.ac.uk} \\\And
  Gabriele Pergola \\
  University of Warwick \\
  \texttt{gabriele.pergola.1@warwick.ac.uk} \\}

\newcommand{\newGispyName}{SciGisPy}

\begin{document}
\maketitle

\begin{abstract}
Biomedical literature is often written in highly specialized language, posing significant comprehension challenges for non-experts. Automatic text simplification (ATS) offers a solution by making such texts more accessible while preserving critical information. However, evaluating ATS for biomedical texts is still challenging due to the limitations of existing evaluation metrics. General-domain metrics like SARI, BLEU, and ROUGE focus on surface-level text features, and readability metrics like FKGL and ARI fail to account for domain-specific terminology or assess how well the simplified text conveys core meanings (\textit{gist}).
To address this, we introduce \textit{\newGispyName}, a novel evaluation metric inspired by Gist Inference Score (GIS) from Fuzzy-Trace Theory (FTT). SciGisPy measures how well a simplified text facilitates the formation of abstract inferences (gist) necessary for comprehension, especially in the biomedical domain. We revise GIS for this purpose by introducing domain-specific enhancements, including semantic chunking, Information Content (IC) theory, and specialized embeddings, while removing unsuitable indices.
Our experimental evaluation on the Cochrane biomedical text simplification dataset demonstrates that \textit{\newGispyName} outperforms the original GIS formulation, with a significant increase in correctly identified simplified texts (84\% versus 44.8\%). The results and a thorough ablation study confirm that \textit{\newGispyName} better captures the essential meaning of biomedical content, outperforming existing approaches.
\end{abstract}

\section{Introduction}
Biomedical literature is often written in highly specialized language, making it challenging for non-experts to understand. The 2022 World Health Organization (WHO) report identifies low public health literacy as a significant global issue, affecting disease prevention and management \cite{WHO}. Automatic text simplification (ATS) offers a potential solution by transforming complex biomedical language into simpler, more accessible text while preserving essential details. However, evaluating the effectiveness of ATS on biomedical texts remains a challenge.

\begin{figure}[t]
  \centering
  \includegraphics[width=\linewidth]{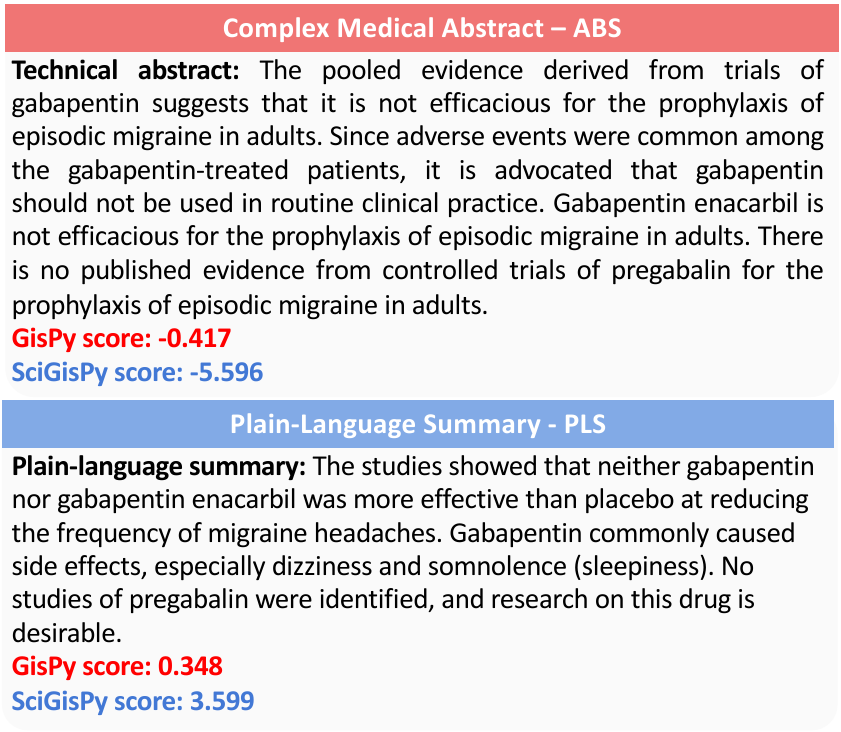}
  \caption{An example of excerpts from a technical abstract (top) and its corresponding plain-language summary (bottom) from the Cochrane text simplification dataset \cite{cochrane_dataset}. \newGispyName\ demonstrates better ability in distinguishing between ABS and PLS.}
  \label{fig:cochrane}
\end{figure}

\begin{figure*}[t]
    \centering
    \subfigure[Original Gist Inference Score (GIS) formula by \cite{gist_original}]{%
        \includegraphics[width=\textwidth]{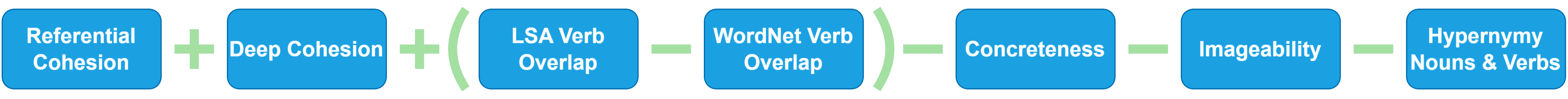}
    }\hfill
    \subfigure[Enhanced GIS formula for Biomedical Text Simplification: \newGispyName]{%
        \includegraphics[width=\textwidth]{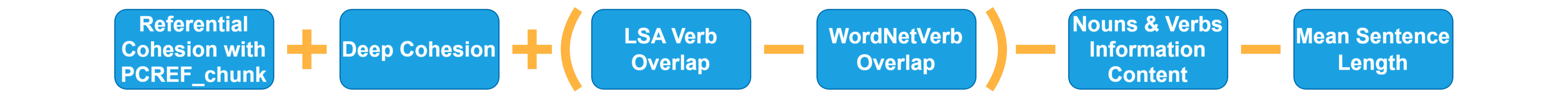}
    }
    \caption{Original and enhanced GIS formula}
    \label{fig:gis_formula}
\end{figure*}

Existing metrics for biomedical text simplification are still limited.  General-domain ATS metrics, such as SARI \cite{SARI}, BLEU \cite{BLEU}, and ROUGE \cite{ROUGE}, focus on surface-level word edits and n-gram overlaps, relying heavily on the quality and variety of reference texts. Similarly, referenceless metrics, like BERTScore \cite{bertscore}, Flesch-Kincaid Grade Level (FKGL) \cite{FKGL}, and Automated Readability Index (ARI) \cite{ARI}, focus on syntactic and lexical simplicity (e.g., shortening sentences, simplifying vocabulary), but fail to capture domain-specific terminology, and more importantly, they cannot ensure that the core meaning (or \textit{gist}) is easily understood. These limitations are especially critical for biomedical texts because they do not provide an adequate measure of whether the text is effective in facilitating comprehension of complex medical concepts despite its linguistic simplicity. 

To address this gap, we propose a novel, task-specific evaluation score, \newGispyName, based on the Gist Inference Score (GIS), inspired by the Fuzzy-Trace Theory (FTT). FTT posits that human cognition operates through two parallel representations: gist (the essential meaning) and verbatim (exact details) \cite{gist_FFT}. GIS measures how effectively a text conveys this \textit{gist}, supporting decision-making. 
While previous GIS formulations have been explored in general domains \cite{hosseini-etal-2022-gispy}, none have been optimized for the complexities of biomedical documents.

In this work, we introduce \newGispyName, the first GIS formulation for biomedical text. Figure \ref{fig:gis_formula} shows the original GIS formula and our enhanced version, which incorporates domain-specific adaptations. These include new indices and improvements, based on semantic-based chunking, Information Content (IC) theory, specialised embeddings, and an overall revision of the original GIS formulation.

\noindent Our contribution can be summarized as follows:
\begin{itemize}
\item We introduce \newGispyName, a novel GIS formulation specifically designed for biomedical text simplification, revising existing indices and eliminating those unsuitable for the biomedical domain. 

\item We introduce newly designed indices for \newGispyName, based on semantic-driven chunking, Information Content (IC) theory, and linguistic features of biomedical sentences.

\item We conduct a comprehensive experimental evaluation of GIS as a metric for biomedical text simplification, analyzing the relevance of each index and its correlation with established simplification metrics.

\end{itemize}

\section{Related Work}

\paragraph{Text Simplification Metrics}
The development of automatic evaluation metrics tailored specifically for biomedical ATS remains under-explored. Due to the scarcity of specialized metrics, existing studies often rely on general-domain ATS metrics, which are insufficient for capturing the characteristics of biomedical text.
Reference-based metrics, such as BLEU \cite{BLEU} and ROUGE \cite{ROUGE}, compare simplified outputs to human-generated references, focusing on n-gram precision and recall. These metrics heavily depend on the quality of reference texts and may penalize valid simplifications that use different wording \cite{pergola19tdam, pergola-etal-2021-disentangled, zhu-etal-2021-topic}. SARI \cite{SARI}, designed for text simplification, evaluates word-level edits but similarly focuses on surface-level features like n-gram overlaps, missing the deeper semantic aspects crucial for biomedical comprehension \cite{sulem-etal-2018-bleu, pergola-etal-2021-boosting, alva-manchego-etal-2021-un}. BERTScore \cite{bertscore}, leveraging contextual embeddings from BERT, improves semantic similarity evaluation but still underperforms in biomedical contexts \cite{sun-etal-2022-phee, zhu-etal-2022-disentangled, zhu-etal-2023-disentangling, lu-etal-2023-napss}.

Readability metrics, such as Flesch-Kincaid Grade Level (FKGL) \cite{FKGL} and Automated Readability Index (ARI) \cite{ARI}, are reference-less and rely on surface features like sentence length and word complexity. However, these metrics do not account for the accurate use of domain-specific terminology or semantic nuances critical to biomedical texts. Consequently, they often fail to effectively evaluate the readability and accuracy of simplified medical content, underscoring the need for more advanced evaluation methods in this domain.

\paragraph{Gist and GIS}
According to Fuzzy-Trace Theory (FTT) \cite{gist_FFT}, individuals encode multiple mental representations when processing text, ranging from verbatim, which captures surface-level details, to gist, which conveys the core meaning. In FTT, "gist" refers to the essential idea of a matter. Prior research \cite{gist_FFT} suggests that gist representations significantly influence decision-making processes more than verbatim representations. Therefore, assessing gist representation can help measure a document’s ability to generate clear and actionable mental models and effectively communicate its message.

The Gist Inference Score (GIS) was first introduced by Wolfe et al. \cite{gist_original} to evaluate how well a text enables readers to form gist inferences. Before the development of the GisPy library \cite{hosseini-etal-2022-gispy}, which automated GIS evaluation, research on GIS was still developing \cite{gist_FFT, gist_original}. GIS was initially proposed by leveraging Coh-Metrix, a multilevel linguistic framework analyzing over 100 variables related to text simplicity, such as referential cohesion, lexical diversity, and latent semantic analysis (LSA) \cite{gist_original, coh-metrix, sun-etal-2024-leveraging}. However, Coh-Metrix lacks batch processing and efficiency.
The GisPy library, building on these earlier methods and leveraging advanced NLP techniques, provides the first open-source solution for computing GIS across multiple documents. In GisPy \cite{hosseini-etal-2022-gispy}, GIS is composed of seven indices: Referential Cohesion, Deep Cohesion, Verb Overlap, Word Concreteness, Word Imageability, and Hypernymy Nouns \& Verbs, each associated with either a positive or negative coefficient. This work extends the GisPy library by modifying, removing, and adding to these indices for better alignment with biomedical simplification tasks.

\subsection{GisPy and Other Text Simplification Metrics}
GIS, as a reference-less metric, is more closely related and comparable to the readability metrics, such as FKGL and ARI. However, we argue that GIS captures different information from the text compared to FKGL and ARI. 

While FKGL and ARI focus on surface-level "verbatim" features of text, GIS aims to measure the likelihood that readers will develop meaningful "gist inferences" from the text. Specifically, FKGL assesses readability based solely on surface-level features using sentence length and word syllable count, whereas GIS captures the underlying abstract meaning by considering more complex dimensions of text features such as cohesion and word concreteness. To validate this argument, we calculated the Pearson correlation coefficients between the reference-less metrics shown in \ref{app: GisPy and Other Text Simplification Metrics}, where the results are all close to zero, proving that GIS is uncorrelated with these metrics.

\section{Method}
In this section, we first review the indices in the original GisPy formulation \cite{hosseini-etal-2022-gispy} and assess their suitability for biomedical text simplification. For each index, we propose adaptations by either (i) introducing novel approaches, (ii) improving the existing indices with more specialized methods, or (iii) removing them if unsuitable for the biomedical domain.

\subsection{Enhancing GIS indices for Biomedical Document Simplification}
As shown in Figure \ref{fig:gis_formula}, the original GIS formula includes seven indices, with some positively and others negatively weighted. These indices cover five dimensions of text features. Through analysis, we posit that only four dimensions are beneficial for evaluating biomedical text simplicity -- \textit{Referential Cohesion, Deep Cohesion, Verb Overlap, Hypernymy Nouns \& Verbs}, while \textit{Word Concreteness and Imageability} is not.

\paragraph{Hypernymy Nouns \& Verbs:} This index measures word specificity, based on the idea that more specific words are harder to understand for a general audience without specialized knowledge. Simplifying biomedical texts often requires translating technical terms into concepts that are accessible to a broader audience, thus making this metric valuable for evaluating simplified biomedical documents.

To achieve this, the index (\texttt{WRDHYPnv}) uses WordNet's hierarchy of concepts and penalizes words with greater depth in the hierarchy, as these represent more specialized terms. In particular, the specificity is quantified by the average hypernym path length of synonym sets.

In the original GIS formula, this index evaluates word specificity by listing all nouns and verbs in a document, identifying their synonym sets in WordNet, and calculating the average hypernym path length. Instead, we propose three more fine-grained alternatives to address limitations when applied to specialized texts: (i) the first ensures proper comparison of noun and verb paths via normalisation (\texttt{WRDHYP\_norm}), (ii) the second introduces and adapts the concept of Information Content \texttt{WRDIC}, and (iii) the third resolves a development issue found in the original GisPy library.

\noindent \underline{\textit{i. Hypernym Root Normalisation:}}
In WordNet, unlike noun synsets that all trace back to the hypernym root '\textit{entity}', verb synsets can trace back to different hypernym roots, with some synsets having multiple roots. For example, in the biomedical domain, the verb \textit{administer} could trace back to \textit{apply} (in the context of giving treatment) or \textit{manage} (in the context of overseeing care). Consequently, the GisPy approach of averaging hypernym paths for all synsets can lead to incomparable path lengths if the roots are different, as these roots may have hypernym hierarchies of varying scales.

To address this issue, we propose an alternative approach, where instead of averaging all hypernym paths, we group the paths that lead to the same root and apply L1 normalization within each group to balance the scales of the hypernym hierarchies. For synsets with multiple hypernym roots, we select the longest hypernym path and its corresponding root as the representative. Finally, we compute the average of the normalized path lengths across all groups to obtain the final result.
We indicate this new index with \texttt{WRDHYP\_norm}, where the suffix stands for ``\textit{root normalization}'', formalised as follows:
\[
\texttt{WRDHYP\_norm} = \frac{1}{n} \sum_{i=1}^{n} \frac{L_i}{||L_i||_1}
\]
Where \( L_i \) is the path length for the \(i\)-th hypernym path group, \( ||L_i||_1 \) is the L1 normalization of the path length for each root group, and \( n \) is the total number of hypernym path groups.

\noindent \underline{\textit{ii. Information Content:}}
To improve the GIS metric for biomedical text simplification, we propose to replace the Wordnet hypernym-based solution \texttt{WRDHYP} with a new approach based on Information Content (IC) \cite{IC}, namely \texttt{WRDIC}. Simple hypernym path counting can be insufficient in some cases, as it fails to account for the frequency and relevance of terms within specific domains. For instance, two biomedical terms may have the same path length but differ significantly in importance or specificity within a corpus. Information Content addresses this issue by considering the probability of encountering a term in a given corpus. In information theory, IC is a measure derived from the probability of a specific event occurring from a random variable \cite{IC}. In this context, the IC of a word can be defined as \(-\log(P(c))\), where \(P(c)\) is the probability of encountering a hypernym of word \(c\) in a corpus.

To compute the new index, similar to the strategy in \texttt{WRDHYP}, we first identify all nouns and verbs in the text. Then, we calculate the average of their IC values to generate the final result. IC provides a more accurate measure of word specificity, with higher IC values indicating more specialized words. This approach enhances the ability to measure text simplification by considering both the structure of the language and its actual use in the domain, making it particularly suitable for specialized domains.

\noindent \underline{\textit{iii. Mean Hypernym Paths Length:}}
The original GisPy score \cite{hosseini-etal-2022-gispy} uses WordNet \cite{wordnet} and its Synset objects from the NLTK library\footnote{\url{https://www.nltk.org}} to compute hierarchical paths (\texttt{WRDHYP}). However, for some verb synsets, there are multiple hypernym roots, resulting in several hypernym paths leading to different roots. This issue is critical because having different roots makes the hypernym paths non-comparable, as the hierarchical structures vary in depth and scope.

The original GisPy paper assumes a single hypernym path per root and does not account for this issue. We addressed this by modifying the index to use the mean length of all available hypernym paths for a given synset when multiple paths are available, which we call \texttt{WRDHYP\_mean} \footnote{We flagged the issue regarding multiple hypernym paths for verb synsets on the GisPy GitHub repository. The authors implemented a solution using the maximum path length, but our preliminary experiments indicated that averaging the path lengths offers a more balanced measure of specificity.}.

\paragraph{Verb Overlap:}
According to the FFT, abstract verb overlaps promote the formation of gist representations, aiding readers in understanding the text's core meaning.
To capture this, the GisPy score uses two indices: \texttt{SMCAUSe} (positively weighted) and \texttt{SMCAUSwn} (negatively weighted) \cite{hosseini-etal-2022-gispy}. For biomedical text simplification, \texttt{SMCAUSe} is important because it promotes simplicity by emphasizing abstract overlaps between verbs, while \texttt{SMCAUSwn} penalizes the redundant repetition of identical or similar verbs.

In its original implementation, this index is based on the \textit{en\_core\_web\_trf}\footnote{\url{https://spacy.io/models/en##en_core_web_trf}} from SpaCy, a RoBERTa-based pre-trained language model \cite{liu2019roberta}, to generate token vector embeddings for each verb, and then computes cosine similarity of the embeddings. 
To better suit the characteristics of biomedical texts, often featuring technical and compound terminology, we propose a simple yet effective modification, adopting embedding models specialised for technical documents. We identify two embedding models for this index, fastText \cite{fasttext} and BioWordVec \cite{biowordvec}.

FastText \cite{fasttext} is a widely used word embedding library, particularly suitable for technical documents. It learns word embeddings on a sub-word basis, which allows it to represent out-of-vocabulary words. This is particularly useful for dealing with biomedical language characterized by many compound words \cite{perg18}. We adopt pre-trained embeddings provided by fastText and name this index \texttt{\textit{SMCAUSf}}.

Our second alternative is BioWordVec \cite{biowordvec}, based on a benchmark biomedical word embedding library. BioWordVec combines subword information from unlabeled biomedical text with the widely-used Biomedical Subject Headings (MeSH) vocabulary \cite{mesh}. Pre-trained using FastText embeddings, BioWordVec is the most commonly used biomedical word embedding model in the recent literature. We adopt it to improve the \texttt{SMCAUS} index, and name it \texttt{SMCAUSb}.

\paragraph{Referential Cohesion:}
Referential Cohesion measures word and idea overlaps across sentences, making it a suitable metric to characterise simplicity in the biomedical text. In \citet{hosseini-etal-2022-gispy} this dimensions is captured with two indices:  \texttt{PCREF} and \texttt{CoREF} (both positively weighted).  \texttt{PCREF}  calculates cosine similarity between sentence embeddings, while the \texttt{CoREF} focuses on coreference resolution across sentences. A high overlap in both indices typically indicates that the text maintains consistent ideas and vocabulary, which helps readers follow complex biomedical content more easily, thus promoting text simplicity. 
To improve the detection of referential cohesion in biomedical texts, we introduce (i) a novel index based on \textit{semantic chunking}, which posits that a lower number of semantic chunks indicates stronger coherence, and (ii) more suitable sentence embedding models designed for technical and biomedical documents.

\noindent \underline{\textit{i. Semantic Chunking:}}
We introduce an alternative solution for measuring Referential Cohesion to substitute \texttt{PCREF}. This new approach is based on the concept of \textit{semantic chunking}\footnote{\href{https://docs.llamaindex.ai/en/stable/examples/node_parsers/semantic_chunking/}{LlamaIndex Semantic Chunking Documentation}}. Unlike traditional methods that chunk text using a fixed size, semantic chunking adaptively determines breakpoints between sentences based on embedding similarity of customizable window size. This ensures that each chunk contains sentences that are semantically related. Similar to \texttt{PCREF}, the semantic chunking method uses cosine similarity between sentences to represent overlap across sentences. 

We argue that a higher number of chunks indicates more diverse semantics and topics within the text. Therefore, minimizing the number of chunks ensures textual coherence and enhances simplicity. Inspired by this, we designed a new index, \texttt{PCREF\_chunk}, built using a semantic chunker to replace the original \texttt{PCREF}. We selected BioSimCSE \cite{biosimcse} as the sentence embedding model, as its biomedical-domain embeddings capture semantics more accurately. We apply a negative coefficient to this index, indicating that fewer semantic chunks correspond to higher coherence and simplicity.

\noindent \underline{\textit{ii. Specialized Sentence Embeddings}}
The original index \textit{PCREF} calculates cosine similarity between sentences using the pretrained MPNet model \footnote{\url{https://huggingface.co/sentence-transformers/all-mpnet-base-v2}} from Hugging Face as the sentence embedding model. 

We experimented with five state-of-the-art sentence embedding models. First, we adopted two leading general-purpose models from the Massive Text Embedding Benchmark (MTEB) Leaderboard mxbai-embed-large-v1 \cite{li2023angle} and the 
e5-mistral-7b-instruct \cite{mistral, wang2023improving, wang2022text} embedding models.
Additionally, we utilized three state-of-the-art biomedical domain embedding models based on BERT: BioSimCSE \cite{biosimcse} and BioBERT \cite{biobert}, which generate contextual embeddings, and a context-free embedding model, BioSentVec \cite{biosentvec}. These models are known for their robustness in biomedical text processing.

Detailed implementation information is provided in Section \ref{sec:experiment}, where a thorough ablation study highlights the impact of each of them. Each model is indicated by a suffix to the index name (e.g., \texttt{PCREF\_mxbai}) specifying the embedding used.

\paragraph{Deep Cohesion:} Indicated by the \textit{PCDC} index (positively weighted), it measures the extent to which a text uses causal and intentional connectives, detected using regular expression patterns. The index is still highly relevant to biomedical text comprehension, as it supports logical relationships between sentences, crucial for ensuring that readers can follow complex biomedical information. Therefore, we retain the original design of this index as in \citet{hosseini-etal-2022-gispy}. 

\paragraph{Mean Sentence Length:} 

FTT suggests that people extract both verbatim (exact details) and gist (core meaning) from texts. Longer or more complex sentences may increase cognitive load, making it harder to focus on gist and potentially promoting reliance on verbatim processing. In contrast, shorter sentences with clear structures could help readers extract gist more easily because the underlying meaning is more accessible. Research in readability and health communication has shown that shorter sentences enhance readability by making information more accessible \cite{rudd2023health}, reducing cognitive load \cite{coh-metrix}, improving comprehension \cite{national2012clear}, and maintaining consistency and focus \cite{weiss2007health}. 

To address this gap within the original GIST score, we propose a new composite index called Mean Sentence Length \texttt{(MSL)} . This index, rewards the reduction of the average sentence length. Concretely, we calculate the mean sentence length by counting the number of words in each sentence and averaging these counts across the entire text. Despite its simplicity, preliminary exploration on this index showed promising results, with a more detailed assessment presented in Section \ref{sec:experiment}.

\subsection{Removing Word Concreteness and Imageability} 
\label{subsec:concrimage} Unlike the previous indices, the dimension we find potentially detrimental to representing biomedical text simplicity is \textit{Word Concreteness and Imageability}. In the original GisPy score, \textit{Word Concreteness} \texttt{(PCCNCz)} measures how concrete and image-evoking words are, while \textit{Imageability} \texttt{(WRDIMGc)} indicates how easily a word can evoke a mental image. For instance, high imageability words like "hammer" are more specific and easily visualized compared to low imageability words like "reason". These indices are negatively weighted in the GIS formula, suggesting its promotion of abstractness. However, we argue that words with high concreteness and imageability (such as “heart”) help understand scientific and biomedical texts, are easier to visualize and thus improve comprehension and make it easier to follow for diverse audiences. Therefore, we hypothesize that removing this concreteness-penalizing index from GIS formula could enhance its performance in biomedical text simplification task.

In conclusion, while the GIS formula promotes abstractness in text to generate Gist, this may not align with promoting simplicity, especially in biomedical texts. Relevant indices may need to be modified or removed to better evaluate simplicity in biomedical documents.

\section{Experiments}
\label{sec:experiment}

\begin{table*}[htb]
    \centering
    \setlength{\tabcolsep}{10pt}
    \resizebox{\textwidth}{!}{
    \begin{tabular}{clcccc}
        \toprule
        \multicolumn{2}{c}{\textbf{Index}} & \textbf{Mean GIS Diff} & \textbf{\% + GIS Diff} & \textbf{\% inc GIS Diff} & \textbf{\% - to +} \\
        \Xhline{0.7pt}
        \multicolumn{2}{c}{Original GisPy \cite{hosseini-etal-2022-gispy}} & -0.461 & 43.38\% & N/A & N/A \\
        \multicolumn{2}{c}{Without \textit{PCCNC} \& \textit{WRDIMG}} & -0.022 & 49.91\% & 60.94\% & \textbf{12.87\%} \\
        \hline
        \multirow{3}{*}{\shortstack{Hypernymy \\ Nouns \& Verbs}} 
        & \textit{WRDHYP\_mean} & 0.158 & 52.85\% & 74.54\% & 11.58\% \\
        & \textit{WRDHYP\_norm} & -1.355 & 31.25\% & 17.56\% & 1.27\% \\
        & \textit{WRDIC} & 0.211 & 51.38\% & \textbf{78.77\%} & 9.74\% \\
        \cline{2-6}
    \multirow{2}{*}{\shortstack{Verb \\ Overlap}} & \textit{SMCAUSe} & -0.728 & 39.25\% & 41.36\% & 3.95\% \\
         & \textit{SMCAUSb} & -0.678 & 40.53\% & 43.38\% & 4.78\% \\
         \cline{2-6}
    \multirow{6}{*}{\shortstack{Referential \\ Cohesion}} & \textit{PCREF\_mxbai} & -0.328 & 45.59\% & 61.12\% & 4.41\% \\
        & \textit{PCREF\_BioSimCSE} & -0.629 & 40.53\% & 37.87\% & 2.48\% \\
        & \textit{PCREF\_BioBERT} & -0.655 & 40.99\% & 43.75\% & 3.68\% \\
        & \textit{PCREF\_mistral} & -0.503 & 44.39\% & 53.22\% & 3.86\% \\
        & \textit{PCREF\_BioSentVec} & -0.686 & 40.99\% & 37.04\% & 2.57\% \\
        & \textit{PCREF\_chunk} & \textbf{0.418} & \textbf{54.32\%} & 61.12\% & 5.97\% \\
        \hline
        \multicolumn{2}{c}{Mean Sentence Length \texttt{(MSL)}} & 0.321 & 52.94\% & 76.93\% & 11.40\% \\
        \hline
        \multicolumn{2}{c}{\textbf{\newGispyName\ (Our)}} & \textbf{2.312} & \textbf{85.39\%} & \textbf{85.57\%} & \textbf{44.21\%} \\
        \Xhline{0.7pt}
    \end{tabular}%
    }
    \caption{Results of GIS Enhancements on Cochrane simplification development set. The second column displays the average GIS difference across the entire dataset. The third column indicates the percentage of documents with a positive GIS difference. The fourth column shows the percentage of documents where the GIS difference increased following the enhancement. The fifth column reports the percentage of documents that initially had a negative GIS difference but shifted to a positive value.}
    \label{tab:gis_results}
\end{table*}

 In this section, we present an experimental evaluation to assess the effectiveness of the GIS metric and our proposed index enhancements in the biomedical domain, using a Cochrane Library dataset \cite{cochrane_dataset} containing technical documents paired with simplified versions, firstly detailed in Section \ref{dataset}. In Section \ref{subsec:experiment_result}, we report the results of our evaluation on this dataset, exploring the impact of different index combinations on simplified texts. Specifically, we analyze which index combinations produce the most significant improvements in gist abstraction by measuring the GIS differences between the technical and simplified documents. Finally, we assess the generalization of our findings by testing on several benchmark datasets from previous GIS literature to evaluate how well our GIS enhancements can be applied beyond the biomedical domain.

\subsection{Datasets}
\label{dataset}
We conducted GIS analysis and tested our indices enhancement on the Cochrane paragraph-level biomedical text simplification dataset \cite{cochrane_dataset}, which is sourced from the Cochrane library\footnote{\url{https://www.cochranelibrary.com/}} of systematic reviews. The Cochrane text simplification dataset comprises 4,459 parallel pairs of technical abstracts (ABS) and their plain-language summaries (PLS) crafted by domain experts, where the PLS texts are simplified versions of original technical abstracts. Figure \ref{fig:cochrane} presents a sample excerpt of a technical abstract and its corresponding PLS. Since GIS score computation does not require any training, we sample 4,334 document pairs as our \textit{development set} to determine the best index configurations; while the remaining additional subset, used as \textit{test set}, will be introduced in the following sections. Since \newGispyName\ does not involve any training, no training set is required.

In previous literature, three benchmark general-domain datasets have been used for evaluating GIS metrics: News Reports vs. Editorials, Journal Article Methods vs. Discussion, and Disneyland Measles Outbreak Data \cite{gist_original, hosseini-etal-2022-gispy}. This subset serves as our \textit{test set} to assess the generalisation of our biomedical-specialized GIS.

\begin{table*}[htb]
    \centering
    \setlength{\tabcolsep}{10pt}
    \resizebox{\textwidth}{!}{%
    \begin{tabular}{ccccc}
        \toprule
        \textbf{Index} & \textbf{Mean GIS Diff} & \textbf{\% + GIS Diff} & \textbf{\% inc GIS Diff} & \textbf{\% - to +} \\
        \hline
        Original GisPy \cite{hosseini-etal-2022-gispy} & -0.311 & 44.8\% & N/A & N/A \\
        \Xhline{0.7pt}
       \textbf{\newGispyName\ (Our)} & 2.295 & 84\% & 79.2\% & 45.6\%  \\
        \Xhline{0.7pt}
    \end{tabular}%
    }
    \caption{Results of GIS Enhancements on Cochrane simplification test set. See Table \ref{tab:gis_results} for column descriptions.}
    \label{tab:gis_results_test}
\end{table*}

\subsection{Results and Discussion}
\label{subsec:experiment_result}
To investigate the effectiveness of applying GisPy GIS in evaluating the simplicity of biomedical text, we first computed GIS values for all Abstracts (ABS)s and Plain Language Summaries (PLS)s in the development set using the best configuration reported for the GisPy library, following the evaluation process outlined in \citet{hosseini-etal-2022-gispy}. After we obtained the GIS score for all documents, we calculated the \textit{GIS difference} between each pair of ABS and PLS:
\[
\text{GIS Difference} = \text{GIS}_{\text{PLS}} - \text{GIS}_{\text{ABS}}
\]

In the rest part of this paper, "GIS difference" refers to the difference calculated using the above equation. A \textit{positive GIS difference} for a pair of documents suggests that audiences will more easily abstract the gist from the simplified text (PLS) compared to the original text (ABS), subsequently showing that GIS can be a good indicator of simplification. Consistent with previous literature, we compare the average GIS difference among all documents under the different GIS formulations to determine the more effective alternatives, namely mean GIS difference.

To evaluate the impact of each individual enhancement, we ran GisPy with each enhancement applied separately, while keeping all other indices identical to those in the original formula. Additionally, when testing combinations of enhancements that modify the same index, for those that modify the same index, we ensured that only one change was applied at a time to prevent overlapping calculations.

\subsubsection{GIS for Biomedical Text Simplification} 
First, we report the GIS scores resulting from the original GisPy on the development set of the Cochrane Simplification Dataset in Table \ref{tab:gis_results}.
The average GIS for ABS texts is 0.225, while for PLS texts, the mean GIS is -0.225, resulting in a mean GIS difference of -0.450; only 43\% of document pairs have a positive GIS difference. These results show that the original GIS formulation struggles to distinguish between simplified and unsimplified texts for most biomedical documents. 

We proceed to discuss the impact of the enhancements proposed in this work; for each index enhancement listed, Table \ref{tab:gis_results} reports the results obtained by running GisPy with only the corresponding enhancements while keeping the rest of the formula unchanged. 

\noindent \textbf{Removing Word Concreteness and Imageability:}
To test our hypothesis that removing word concreteness and imageability promotes biomedical text easier to comprehend (Sec. \ref{subsec:concrimage}), we ran GIS without \textit{PCCNC} and \textit{WRDIMG} and observed positive results, as reported in Table \ref{tab:gis_results}: the average GIS difference increased from -0.450 to -0.022, with \textbf{49.91\%} of documents now exhibiting a positive GIS difference, compared to 43.38\% with the original GIS formula. This finding supports our initial analysis and confirms the need to tailor the roles of the indices when dealing with specialised domains.

\noindent \textbf{Semantic Chunking:} As mentioned earlier, the mean GIS difference is our primary metric for evaluating the performance of new GIS formula. A larger difference indicates better distinction between simplified and original documents. Based on the experiment results shown in Table \ref{tab:gis_results}, most of our enhancements produced positive outcomes. The enhancement \textit{PCREF\_chunk} achieved the most significant improvement, leading to the mean GIS difference increase from -0.461 to 0.418. This enhancement also led to 54.32\% of documents obtaining a positive GIS difference, an increase of 10.94\% compared to the original GisPy, which achieved 43.38\%.

In addition, we tracked the impact of each enhancement on individual documents. The fourth column in Table \ref{tab:gis_results} presents the percentage of documents where the GIS difference increased after the enhancement. This indicates that the enhanced GIS formula can better distinguish between the original ABS text and the simplified PLS text compared to the original GIS. The table's last column also shows the percentage of documents that originally had a negative GIS difference but switched to positive; this represents cases where the original GIS failed to evaluate simplicity, but the new GIS succeeded.

Looking at ABS-PLS pairs in Table \ref{tab:gis_results}, more than half of our indices enhancements yielded positive results. Some indices demonstrated significant improvements, with \textit{WRDIC} by achieving the highest increase with 78.77\% of documents in the development set transiting to a positive GIS difference.

\noindent \textbf{Best Formulation:} Based on the experimental results on the development set, we identified the best combination of our enhanced GIS formula, as shown in Figure \ref{fig:gis_formula}. We adopted the enhancements of Referential Cohesion with Semantic Chunking (\texttt{PCREF\_chunk}), Hypernyms with Information Content (IC) (\texttt{WRDIC}), and Mean Sentence Length (\texttt{MSL}), together with the removal of indices \texttt{PCCNC} and \texttt{WRDIMG}. The significant results of this biomedical text simplification-targeted GIS formula are presented in the last row of Table \ref{tab:gis_results}. 

\noindent \textbf{Generalisation:} To test the generalisation of this finding, we also applied the enhanced formula to the Cochrane test set. The results, presented in the last row of Table \ref{tab:gis_results_test}, demonstrate a significant improvement, with the new GIS successfully identifying 84\% of simplified texts, doubling the original number. This confirms the effectiveness of our new GIS for evaluating biomedical text simplification.

\subsubsection{Gist Inference Benchmarks} 
\begin{table*}[ht]
    \centering
    \setlength{\tabcolsep}{10pt}
    \resizebox{0.99\textwidth}{!}{%
    \begin{tabular}{l l c c c}
        \toprule
        \textbf{Benchmark} & \textbf{Approach} & \textbf{Distance} & \textbf{t-statistic} & \textbf{p-value} \\
        \Xhline{0.7pt}
        \multirow{4}{*}{Reports vs. Editorials} 
        & GisPy with \textit{PCREF\_mistral} \& \textit{MSL}&  \textbf{3.260} & 4.068 & * $2 \times 10^{-4}$ \\
        & GisPy \cite{hosseini-etal-2022-gispy} &  2.551 & 3.643 & * $7 \times 10^{-4}$ \\
        & Coh-Metrix \cite{coh-metrix} & 2.535 & 3.826 & * $3 \times 10^{-4}$ \\
        & \cite{gist_original} & 0.368 & - & - \\
        \hline
        \multirow{4}{*}{Methods vs. Discussion} 
        & GisPy with \textit{SCAUSf} & \textbf{5.200} & 5.916 & * $3 \times 10^{-7}$ \\
        & GisPy  & 5.012 & 7.188 & * $3 \times 10^{-9}$ \\
        & Coh-Metrix  & 5.010 & 6.331 & * $7 \times 10^{-8}$ \\
        & \cite{gist_original}  & 0.747 & - & - \\
        \hline
        \multirow{3}{*}{Disney} 
        & GisPy with \textit{MSL} & \textbf{2.442} & 3.492 & * $6 \times 10^{-4}$ \\
        & GisPy & 2.418 & 3.440 & * $7 \times 10^{-4}$ \\
        & Coh-Metrix  & 0.998 & 1.878 & $6 \times 10^{-2}$ \\
        \bottomrule
    \end{tabular}
    }
    \caption{Comparison of GIS scores generated by GisPy with our enhancement indices vs. original GisPy vs. other methods for all benchmarks}
    \label{tab:gis_comparison}
\end{table*}
To assess whether our enhancements improve the evaluation of Gist abstraction in the general domain, the original objective of GIS, we tested all index enhancements on the benchmark datasets used in the original GisPy paper. 
The News Reports vs. Editorials dataset comprises 50 pairs of documents per category, totaling 100 documents. The Journal Article Methods vs. Discussion dataset includes 25 pairs, amounting to 50 documents. The Disneyland dataset consists of 191 articles in total. To ensure comparability with these datasets, we randomly sampled 125 document pairs from the Cochrane dataset.

The experimental results were less significant compared to the previous results on the Cochrane simplification dataset since our enhancements were targeted at biomedical text simplification. However, we still identified a combination of index enhancements that outperformed the original GisPy formula on the benchmark dataset. The results are presented in Table \ref{tab:gis_comparison}, where we also performed a student’s t-test with the null hypothesis following GisPy \cite{hosseini-etal-2022-gispy} paper, which shows how good a GIS score can significantly distinguish these ABS texts and PLS texts. This positive result demonstrates that our proposed solutions are not only beneficial for simplification evaluation, but also enhance the measure of how easily GIS can be inferred.

\section{Conclusion}
In this study, we addressed the challenge of evaluating biomedical automatic text simplification by introducing a novel referenceless evaluation metric, \newGispyName, inspired by the Gist Inference Score (GIS) from Fuzzy-Trace Theory. This metric was specifically adapted and enhanced for biomedical text simplification through rigorous feasibility analysis and domain-specific enhancements. Our comprehensive experimental assessment on the Cochrane text simplification dataset demonstrates that \newGispyName\ significantly outperforms the original GIS metric in assessing the simplicity of biomedical texts.

\section{Limitations}
A limitation of this study is the reliance on a single benchmark, the Cochrane simplification dataset, due to the limited availability and suitability of biomedical text simplification datasets at the document level. Validating our methodology across multiple datasets would strengthen its robustness. 

Additionally, while we introduced several improvements to the individual GIS indices, the coefficient magnitudes currently remain fixed at 1. Developing an automated method to dynamically adjust these coefficients based on text distributions could further improve the accuracy and versatility of \newGispyName\ in text simplification.

\section*{Acknowledgements}
This work was supported by Sulis, the Tier-2 high-performance computing facility at the University of Warwick, funded by the EPSRC through the HPC Midlands+ Consortium (EPSRC grant no. EP/T022108/1). 



\bibliography{custom}

\begin{thebibliography}{40}
\providecommand{\natexlab}[1]{#1}

\bibitem[{Alva-Manchego et~al.(2021)Alva-Manchego, Scarton, and Specia}]{alva-manchego-etal-2021-un}
Fernando Alva-Manchego, Carolina Scarton, and Lucia Specia. 2021.
\newblock \href {https://doi.org/10.1162/coli_a_00418} {The (un)suitability of automatic evaluation metrics for text simplification}.
\newblock \emph{Computational Linguistics}, 47(4):861--889.

\bibitem[{Bojanowski et~al.(2017)Bojanowski, Grave, Joulin, and Mikolov}]{fasttext}
Piotr Bojanowski, Edouard Grave, Armand Joulin, and Tomas Mikolov. 2017.
\newblock Enriching word vectors with subword information.
\newblock \emph{Transactions of the association for computational linguistics}, 5:135--146.

\bibitem[{Chen et~al.(2019)Chen, Peng, and Lu}]{biosentvec}
Qingyu Chen, Yifan Peng, and Zhiyong Lu. 2019.
\newblock Biosentvec: creating sentence embeddings for biomedical texts.
\newblock In \emph{2019 IEEE International Conference on Healthcare Informatics (ICHI)}, pages 1--5. IEEE.

\bibitem[{Cover and Thomas(2006)}]{IC}
Thomas~M. Cover and Joy~A. Thomas. 2006.
\newblock \emph{Elements of Information Theory (Wiley Series in Telecommunications and Signal Processing)}.
\newblock Wiley-Interscience, USA.

\bibitem[{Devaraj et~al.(2021)Devaraj, Wallace, Marshall, and Li}]{cochrane_dataset}
Ashwin Devaraj, Byron~C Wallace, Iain~J Marshall, and Junyi~Jessy Li. 2021.
\newblock Paragraph-level simplification of medical texts.
\newblock In \emph{Proceedings of the conference. Association for Computational Linguistics. North American Chapter. Meeting}, volume 2021, page 4972. NIH Public Access.

\bibitem[{Graesser et~al.(2011)Graesser, McNamara, and Kulikowich}]{coh-metrix}
Arthur~C Graesser, Danielle~S McNamara, and Jonna~M Kulikowich. 2011.
\newblock Coh-metrix: Providing multilevel analyses of text characteristics.
\newblock \emph{Educational researcher}, 40(5):223--234.

\bibitem[{Hosseini et~al.(2022)Hosseini, Wolfe, Diab, and Broniatowski}]{hosseini-etal-2022-gispy}
Pedram Hosseini, Christopher Wolfe, Mona Diab, and David Broniatowski. 2022.
\newblock \href {https://doi.org/10.18653/v1/2022.wnu-1.5} {{G}is{P}y: A tool for measuring gist inference score in text}.
\newblock In \emph{Proceedings of the 4th Workshop of Narrative Understanding (WNU2022)}, pages 38--46, Seattle, United States. Association for Computational Linguistics.

\bibitem[{Jiang et~al.(2023)Jiang, Sablayrolles, Mensch, Bamford, Chaplot, Casas, Bressand, Lengyel, Lample, Saulnier et~al.}]{mistral}
Albert~Q Jiang, Alexandre Sablayrolles, Arthur Mensch, Chris Bamford, Devendra~Singh Chaplot, Diego de~las Casas, Florian Bressand, Gianna Lengyel, Guillaume Lample, Lucile Saulnier, et~al. 2023.
\newblock Mistral 7b.
\newblock \emph{arXiv preprint arXiv:2310.06825}.

\bibitem[{Kanakarajan et~al.(2022)Kanakarajan, Kundumani, Abraham, and Sankarasubbu}]{biosimcse}
Kamal~Raj Kanakarajan, Bhuvana Kundumani, Abhijith Abraham, and Malaikannan Sankarasubbu. 2022.
\newblock Biosimcse: Biomedical sentence embeddings using contrastive learning.
\newblock In \emph{Proceedings of the 13th International Workshop on Health Text Mining and Information Analysis (LOUHI)}, pages 81--86.

\bibitem[{Kincaid et~al.(1975)Kincaid, Fishburne, Rogers, and Chissom}]{FKGL}
Peter Kincaid, Robert~P. Fishburne, Richard~L. Rogers, and Brad~S. Chissom. 1975.
\newblock \href {https://api.semanticscholar.org/CorpusID:61131325} {Derivation of new readability formulas (automated readability index, fog count and flesch reading ease formula) for navy enlisted personnel}.

\bibitem[{Lee et~al.(2020)Lee, Yoon, Kim, Kim, Kim, So, and Kang}]{biobert}
Jinhyuk Lee, Wonjin Yoon, Sungdong Kim, Donghyeon Kim, Sunkyu Kim, Chan~Ho So, and Jaewoo Kang. 2020.
\newblock Biobert: a pre-trained biomedical language representation model for biomedical text mining.
\newblock \emph{Bioinformatics}, 36(4):1234--1240.

\bibitem[{Li and Li(2023)}]{li2023angle}
Xianming Li and Jing Li. 2023.
\newblock Angle-optimized text embeddings.
\newblock \emph{arXiv preprint arXiv:2309.12871}.

\bibitem[{Lin(2004)}]{ROUGE}
Chin-Yew Lin. 2004.
\newblock Rouge: A package for automatic evaluation of summaries.
\newblock In \emph{Text summarization branches out}, pages 74--81.

\bibitem[{Lipscomb(2000)}]{mesh}
C.E. Lipscomb. 2000.
\newblock {Medical Subject Headings (MeSH)}.
\newblock \emph{Bulletin of the Medical Library Association}, 88(3):265.

\bibitem[{Liu et~al.(2021)Liu, Lin, Shi, and Zhao}]{liu2019roberta}
Zhuang Liu, Wayne Lin, Ya~Shi, and Jun Zhao. 2021.
\newblock A robustly optimized bert pre-training approach with post-training.
\newblock In \emph{China National Conference on Chinese Computational Linguistics}, pages 471--484. Springer.

\bibitem[{Lu et~al.(2023)Lu, Li, Wallace, He, and Pergola}]{lu-etal-2023-napss}
Junru Lu, Jiazheng Li, Byron Wallace, Yulan He, and Gabriele Pergola. 2023.
\newblock \href {https://doi.org/10.18653/v1/2023.findings-eacl.80} {{N}ap{SS}: Paragraph-level medical text simplification via narrative prompting and sentence-matching summarization}.
\newblock In \emph{Findings of the Association for Computational Linguistics: EACL 2023}, pages 1079--1091, Dubrovnik, Croatia. Association for Computational Linguistics.

\bibitem[{Miller(1995)}]{wordnet}
George~A Miller. 1995.
\newblock Wordnet: a lexical database for english.
\newblock \emph{Communications of the ACM}, 38(11):39--41.

\bibitem[{{National Institutes of Health}(2012)}]{national2012clear}
{National Institutes of Health}. 2012.
\newblock Clear communication: An {NIH} health literacy initiative.
\newblock \emph{http://www. nih. gov/clearcommunication/healtHealth Literacyiteracy. htm}.

\bibitem[{Osborne et~al.(2022)Osborne, Elmer, Hawkins, Cheng, Batterham, Dias, Good, Monteiro, Mikkelsen, Nadarajah, and Fones}]{WHO}
Richard Osborne, Shandell Elmer, Melanie Hawkins, Christina Cheng, Roy Batterham, Sónia Dias, Suvajee Good, Maristela Monteiro, Bente Mikkelsen, Ranjit Nadarajah, and Guy Fones. 2022.
\newblock Health literacy development is central to the prevention and control of non-communicable diseases.
\newblock \emph{BMJ Global Health}, 7.

\bibitem[{Papineni et~al.(2002)Papineni, Roukos, Ward, and Zhu}]{BLEU}
Kishore Papineni, Salim Roukos, Todd Ward, and Wei-Jing Zhu. 2002.
\newblock Bleu: a method for automatic evaluation of machine translation.
\newblock In \emph{Proceedings of the 40th annual meeting of the Association for Computational Linguistics}, pages 311--318.

\bibitem[{Pergola et~al.(2019)Pergola, Gui, and He}]{pergola19tdam}
Gabriele Pergola, Lin Gui, and Yulan He. 2019.
\newblock {TDAM}: A topic-dependent attention model for sentiment analysis.
\newblock \emph{Information Processing \& Management}, 56(6):102084.

\bibitem[{Pergola et~al.(2021{\natexlab{a}})Pergola, Gui, and He}]{pergola-etal-2021-disentangled}
Gabriele Pergola, Lin Gui, and Yulan He. 2021{\natexlab{a}}.
\newblock \href {https://doi.org/10.18653/v1/2021.naacl-main.228} {A disentangled adversarial neural topic model for separating opinions from plots in user reviews}.
\newblock In \emph{Proceedings of the 2021 Conference of the North American Chapter of the Association for Computational Linguistics: Human Language Technologies}, pages 2870--2883, Online. Association for Computational Linguistics.

\bibitem[{Pergola et~al.(2018)Pergola, He, and Lowe}]{perg18}
Gabriele Pergola, Yulan He, and David Lowe. 2018.
\newblock Topical phrase extraction from clinical reports by incorporating both local and global context.
\newblock In \emph{The 2nd AAAI Workshop on Health Intelligence}, pages 499--506.
\newblock Association for the Advancement of Artificial Intelligence. 2018 Workshop on Health Intelligence (W3PHIAI 2018).

\bibitem[{Pergola et~al.(2021{\natexlab{b}})Pergola, Kochkina, Gui, Liakata, and He}]{pergola-etal-2021-boosting}
Gabriele Pergola, Elena Kochkina, Lin Gui, Maria Liakata, and Yulan He. 2021{\natexlab{b}}.
\newblock \href {https://doi.org/10.18653/v1/2021.eacl-main.169} {Boosting low-resource biomedical {QA} via entity-aware masking strategies}.
\newblock In \emph{Proceedings of the 16th Conference of the European Chapter of the Association for Computational Linguistics: Main Volume}, pages 1977--1985, Online. Association for Computational Linguistics.

\bibitem[{Reyna(2021)}]{gist_FFT}
Valerie~F Reyna. 2021.
\newblock A scientific theory of gist communication and misinformation resistance, with implications for health, education, and policy.
\newblock \emph{Proceedings of the National Academy of Sciences}, 118(15):e1912441117.

\bibitem[{Rudd et~al.(2023)Rudd, Anderson, Oppenheimer, and Nath}]{rudd2023health}
Rima~E Rudd, Jennie~Epstein Anderson, Sarah Oppenheimer, and Charlotte Nath. 2023.
\newblock Health literacy: an update of medical and public health literature.
\newblock In \emph{Review of Adult Learning and Literacy, Volume 7}, pages 175--204. Routledge.

\bibitem[{Senter and Smith(1967)}]{ARI}
RJ~Senter and Edgar~A Smith. 1967.
\newblock Automated readability index.
\newblock Technical report, Technical report, DTIC document.

\bibitem[{Sulem et~al.(2018)Sulem, Abend, and Rappoport}]{sulem-etal-2018-bleu}
Elior Sulem, Omri Abend, and Ari Rappoport. 2018.
\newblock \href {https://doi.org/10.18653/v1/D18-1081} {{BLEU} is not suitable for the evaluation of text simplification}.
\newblock In \emph{Proceedings of the 2018 Conference on Empirical Methods in Natural Language Processing}, pages 738--744, Brussels, Belgium. Association for Computational Linguistics.

\bibitem[{Sun et~al.(2022)Sun, Li, Pergola, Wallace, John, Greene, Kim, and He}]{sun-etal-2022-phee}
Zhaoyue Sun, Jiazheng Li, Gabriele Pergola, Byron Wallace, Bino John, Nigel Greene, Joseph Kim, and Yulan He. 2022.
\newblock \href {https://doi.org/10.18653/v1/2022.emnlp-main.376} {{PHEE}: A dataset for pharmacovigilance event extraction from text}.
\newblock In \emph{Proceedings of the 2022 Conference on Empirical Methods in Natural Language Processing}, pages 5571--5587, Abu Dhabi, United Arab Emirates. Association for Computational Linguistics.

\bibitem[{Sun et~al.(2024)Sun, Pergola, Wallace, and He}]{sun-etal-2024-leveraging}
Zhaoyue Sun, Gabriele Pergola, Byron Wallace, and Yulan He. 2024.
\newblock \href {https://aclanthology.org/2024.eacl-short.30} {Leveraging {C}hat{GPT} in pharmacovigilance event extraction: An empirical study}.
\newblock In \emph{Proceedings of the 18th Conference of the European Chapter of the Association for Computational Linguistics (Volume 2: Short Papers)}, pages 344--357, St. Julian{'}s, Malta. Association for Computational Linguistics.

\bibitem[{Wang et~al.(2022)Wang, Yang, Huang, Jiao, Yang, Jiang, Majumder, and Wei}]{wang2022text}
Liang Wang, Nan Yang, Xiaolong Huang, Binxing Jiao, Linjun Yang, Daxin Jiang, Rangan Majumder, and Furu Wei. 2022.
\newblock Text embeddings by weakly-supervised contrastive pre-training.
\newblock \emph{arXiv preprint arXiv:2212.03533}.

\bibitem[{Wang et~al.(2023)Wang, Yang, Huang, Yang, Majumder, and Wei}]{wang2023improving}
Liang Wang, Nan Yang, Xiaolong Huang, Linjun Yang, Rangan Majumder, and Furu Wei. 2023.
\newblock Improving text embeddings with large language models.
\newblock \emph{arXiv preprint arXiv:2401.00368}.

\bibitem[{Weiss(2007)}]{weiss2007health}
Barry~D Weiss. 2007.
\newblock \emph{Health literacy and patient safety: Help patients understand. Manual for clinicians}.
\newblock American Medical Association Foundation.

\bibitem[{Wolfe et~al.(2019)Wolfe, Dandignac, and Reyna}]{gist_original}
Christopher~R Wolfe, Mitchell Dandignac, and Valerie~F Reyna. 2019.
\newblock A theoretically motivated method for automatically evaluating texts for gist inferences.
\newblock \emph{Behavior research methods}, 51:2419--2437.

\bibitem[{Xu et~al.(2016)Xu, Napoles, Pavlick, Chen, and Callison-Burch}]{SARI}
Wei Xu, Courtney Napoles, Ellie Pavlick, Quanze Chen, and Chris Callison-Burch. 2016.
\newblock Optimizing statistical machine translation for text simplification.
\newblock \emph{Transactions of the Association for Computational Linguistics}, 4:401--415.

\bibitem[{Zhang et~al.(2019{\natexlab{a}})Zhang, Kishore, Wu, Weinberger, and Artzi}]{bertscore}
Tianyi Zhang, Varsha Kishore, Felix Wu, Kilian~Q Weinberger, and Yoav Artzi. 2019{\natexlab{a}}.
\newblock Bertscore: Evaluating text generation with bert.
\newblock \emph{arXiv preprint arXiv:1904.09675}.

\bibitem[{Zhang et~al.(2019{\natexlab{b}})Zhang, Chen, Yang, Lin, and Lu}]{biowordvec}
Yijia Zhang, Qingyu Chen, Zhihao Yang, Hongfei Lin, and Zhiyong Lu. 2019{\natexlab{b}}.
\newblock Biowordvec, improving biomedical word embeddings with subword information and mesh.
\newblock \emph{Scientific data}, 6(1):52.

\bibitem[{Zhu et~al.(2022)Zhu, Fang, Pergola, Procter, and He}]{zhu-etal-2022-disentangled}
Lixing Zhu, Zheng Fang, Gabriele Pergola, Robert Procter, and Yulan He. 2022.
\newblock \href {https://doi.org/10.18653/v1/2022.naacl-main.112} {Disentangled learning of stance and aspect topics for vaccine attitude detection in social media}.
\newblock In \emph{Proceedings of the 2022 Conference of the North American Chapter of the Association for Computational Linguistics: Human Language Technologies}, pages 1566--1580, Seattle, United States. Association for Computational Linguistics.

\bibitem[{Zhu et~al.(2021)Zhu, Pergola, Gui, Zhou, and He}]{zhu-etal-2021-topic}
Lixing Zhu, Gabriele Pergola, Lin Gui, Deyu Zhou, and Yulan He. 2021.
\newblock \href {https://doi.org/10.18653/v1/2021.acl-long.125} {Topic-driven and knowledge-aware transformer for dialogue emotion detection}.
\newblock In \emph{Proceedings of the 59th Annual Meeting of the Association for Computational Linguistics and the 11th International Joint Conference on Natural Language Processing (Volume 1: Long Papers)}, pages 1571--1582, Online. Association for Computational Linguistics.

\bibitem[{Zhu et~al.(2023)Zhu, Zhao, Pergola, and He}]{zhu-etal-2023-disentangling}
Lixing Zhu, Runcong Zhao, Gabriele Pergola, and Yulan He. 2023.
\newblock \href {https://doi.org/10.18653/v1/2023.findings-acl.115} {Disentangling aspect and stance via a {S}iamese autoencoder for aspect clustering of vaccination opinions}.
\newblock In \emph{Findings of the Association for Computational Linguistics: ACL 2023}, pages 1827--1842, Toronto, Canada. Association for Computational Linguistics.

\end{thebibliography}

\newpage

\appendix

\setcounter{table}{0}
\renewcommand{\thetable}{A\arabic{table}}
\setcounter{figure}{0}
\renewcommand{\thefigure}{A\arabic{figure}}

\section{Appendix}
\subsection{Original GIS distribution on Cochrane dataset}
\begin{figure}[hb]
    \centering
    \subfigure[GIS distribution for ABS]{%
        \includegraphics[width=0.45\textwidth]{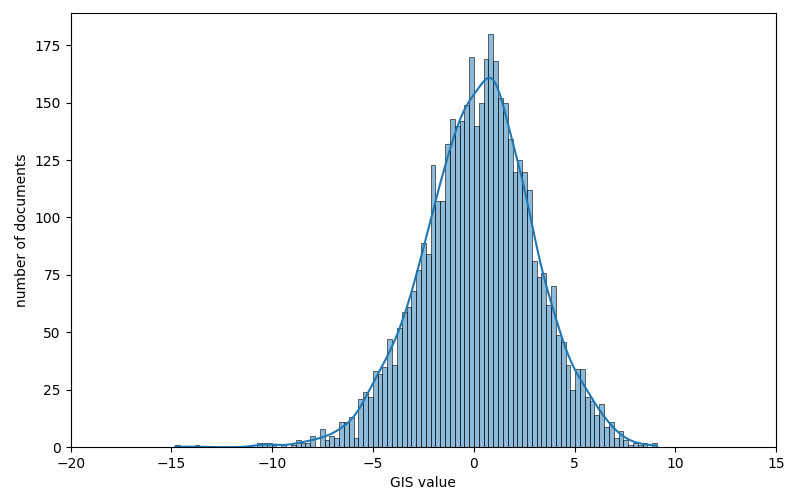}
    }\hfill
    \subfigure[GIS distribution for PLS]{%
        \includegraphics[width=0.45\textwidth]{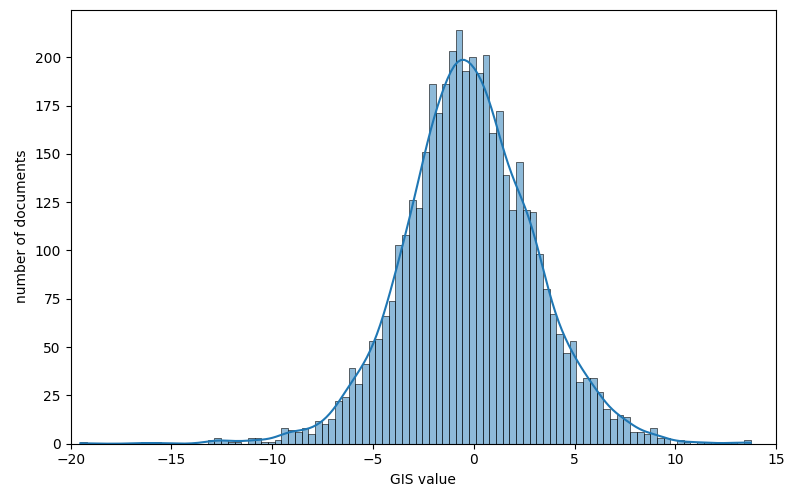}
    }
    \caption{GIS distribution histogram and KDE for Cochrane text simplification dataset}
    \label{fig:gis_dist}
\end{figure}

The GIS distributions of ABS and PLS are jointly shown in Figure \ref{fig:gis_dist}. Both distributions resemble Gaussian distributions, since all indices in GIS were transformed into z-scores, which were subsequently summed up with coefficients to GIS.

\subsection{GIS correlation with other TS metrics}
\begin{figure}[ht]
    \centering
    \subfigure[GIS vs. FKGL for ABS]{\includegraphics[width=0.23\textwidth]{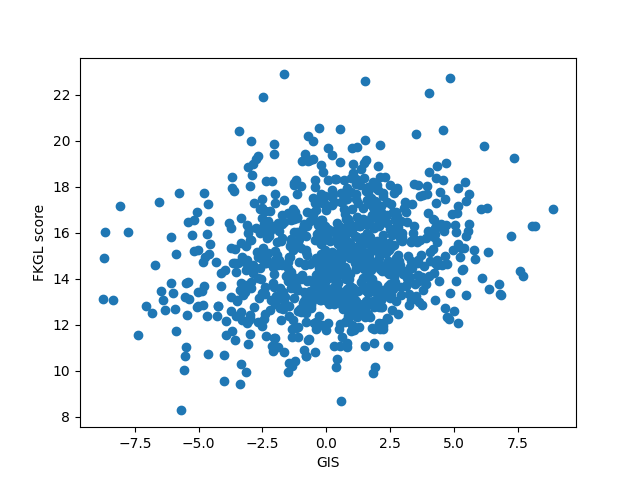}}
    \subfigure[GIS vs. FKGL for PLS]{\includegraphics[width=0.23\textwidth]{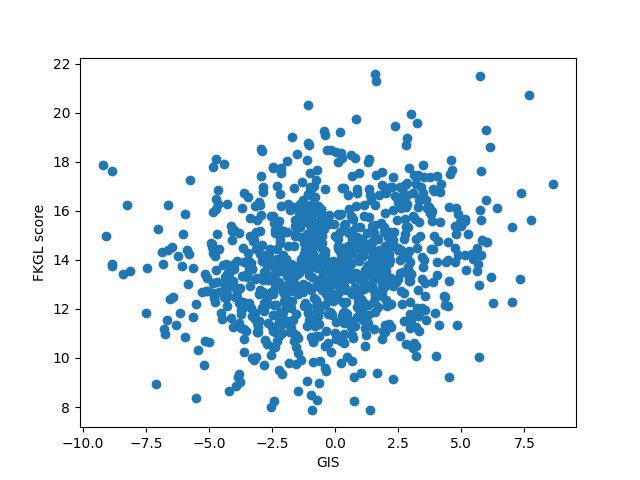}}
    \subfigure[GIS vs. ARI for ABS]{\includegraphics[width=0.23\textwidth]{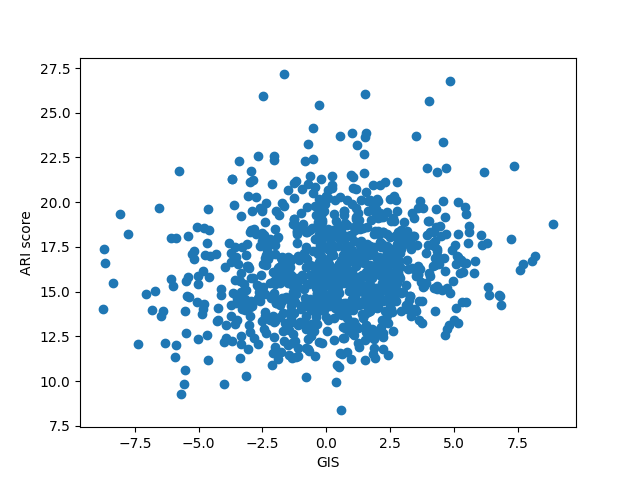}}
    \subfigure[GIS vs. ARI for PLS]{\includegraphics[width=0.23\textwidth]{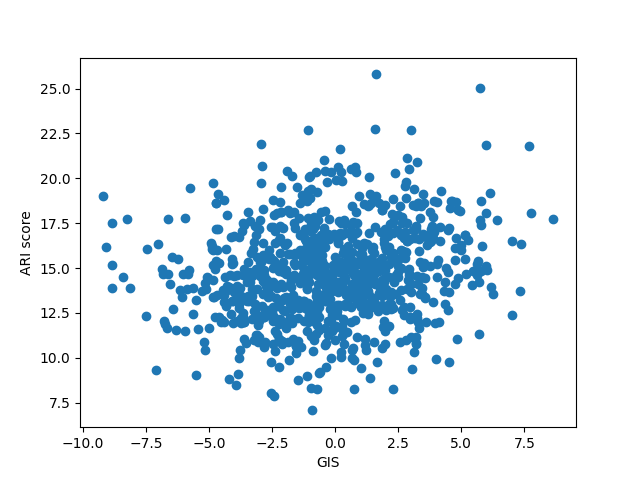}}
    \caption{Scatter plot for GIS and other metrics on Cochrane simplification dataset, on ABS and PLS documents separately}
    \label{fig:scatter}
\end{figure}

This is initially illustrated in Figure \ref{fig:scatter}, where ABS and PLS from a subset of Cochrane simplification dataset (1000 samples) are plotted on corresponding scatter plots, with GIS on the vertical axis and the respective text simplification metric on the horizontal axis. Here we sampled 1000 documents from the development set due to the difficulty to visualize the original large amount of data. If there were a correlation, the points would roughly form a line, however this is not observed in any of the plots.

\subsection{GisPy and Other Text Simplification Metrics}
\label{app: GisPy and Other Text Simplification Metrics}
In this section, we demonstrate that there is no overlap between the aspects evaluated by GIS and other automatic text simplification metrics (FKGL and ARI), with highlighting the unique advantages of using GIS for this task. 

\begin{table}[h!]
    \centering
    \small
    \begin{tabular}{c|cc}
        \toprule
        \textbf{Documents} & \textbf{GIS vs. FKGL} & \textbf{GIS vs. ARI} \\
        \hline
        ABS & 0.17 & 0.14 \\
        
        PLS & 0.18 & 0.18 \\
        \bottomrule
    \end{tabular}
    \caption{Pearson correlation coefficients between GIS and FKGL, and between GIS and ARI}
    \label{tab:pearson}
\end{table}

To validate the above argument, we calculated the Pearson correlation coefficients between the reference-less metrics shown in \ref{tab:pearson}, where the results are all close to zero: for between GIS and KFGL, the numbers are 0.17 for ABS texts and 0.18 for PLS texts; for between GIS and ARI, the coefficient is 0.14 for ABS texts, and 0.18 for PLS texts. Note that the Pearson correlation coefficient would suggest no linear correlation if the value between two distributions is close to 0. These results further prove that GIS is uncorrelated with these metrics.

\end{document}